\documentclass[10pt, a4paper]{article}

\usepackage{lrec2026} 
\usepackage{amssymb}
\usepackage{pifont}
\usepackage{amsmath}
\usepackage{xcolor}
\usepackage{booktabs}

\title{Speak in Context: Multilingual ASR with Speech-Context Alignment via Contrastive Learning}

\name{Yuchen Zhang\textsuperscript{1,2}, Haralambos Mouratidis\textsuperscript{1,2}, Ravi Shekhar\textsuperscript{1,2} } 

\address{\textsuperscript{1}Institute for Analytics and Data Science, University of Essex \\
\textsuperscript{2}School of Computer Science and Electronic Engineering, University of Essex \\  
\texttt{\{yuchen.zhang,r.shekhar,h.mouratidis\}@essex.ac.uk}}

\abstract{
Automatic speech recognition (ASR) has benefited from advances in pretrained speech and language models, yet most systems remain constrained to monolingual settings and short, isolated utterances. While recent efforts in context-aware ASR show promise, two key challenges persist: limited multilingual support and the absence of principled alignment between speech and contextual representations.
In this paper, we introduce a context-aware multilingual ASR framework that supports diverse languages and accents while preserving the modularity of pretrained models. Our approach combines a frozen speech encoder and a decoder-only language model via a lightweight projection module, allowing structured context prompts, including dialogue history and biasing words, to guide transcription. To improve interaction between speech and context, we employ a contrastive learning objective that aligns their representations in a shared embedding space. Evaluations on over 1,500 hours of real-world conversational speech across 11 languages and 5 English dialects show that contextual input consistently improves recognition quality. Contrastive alignment provides additional gains when applied to different context types, with an overall performance gain of over 5\%. These results highlight the importance of both contextual modeling and cross-modal alignment in multilingual ASR. 
\\ \newline  }

\begin{document}

\maketitleabstract

\section{Introduction}

Automatic speech recognition (ASR) has advanced rapidly in recent years, largely due to the development of large-scale pretrained models and end-to-end architectures. However, real-world ASR systems still face persistent challenges in multilingual scenarios. 
Recent developments in speech‑language model integration have paved new ways to connect pretrained speech encoders with large language models (LLMs), enabling speech‑to‑text transcription via prompt‑based mechanisms while often keeping backbone components frozen \citep{verdini2024connect}. For example, \citet{hono2024integrating} proposes integrating a pre‑trained speech representation model with an LLM through a bridge network, achieving competitive end‑to‑end ASR performance.  \citet{fathullah2024prompting} demonstrates that attaching a small audio encoder to a frozen LLM allows multilingual speech recognition, even though the LLM was trained primarily on English text.  

However, two critical gaps remain. First, multilingual context‑aware ASR, which supports multiple languages and actively integrates preceding conversational context or biasing lists, has been relatively understudied. For instance, \citet{cheng2024context} investigates prompt‑based context‑aware recognition in accented speech but focuses on monolingual short‑utterance scenarios rather than fully multilingual conversational settings.  Second, while many works incorporate context as additional input such as concatenating previous utterances or bias lists \citep{zufle2024contrastive, yang2024ctc} , the explicit alignment between speech embeddings and contextual embeddings via a trainable, embedding‑level modality alignment mechanism remains largely unexplored. For example, \citet{guo2021context} introduces a context‑aware language model that encodes callsign lists alongside ASR decoding but does not explore embedding‑space alignment between acoustic and context representations. 

To address these gaps, we propose a context-aware multilingual ASR framework  \footnote{The code is available at \url{https://github.com/yuchen-zhang-essex/Context-Aware_ASR}.} that supports cross-lingual recognition while explicitly aligning speech and contextual representations in the embedding space, moving beyond heuristic concatenation. Our method integrates a frozen speech encoder with a frozen decoder-only LLM through a lightweight projection module, and incorporates structured context information into the LLM input. To enhance cross-modal alignment, we introduce a contrastive learning objective that draws speech–context pairs closer in the shared representation space. This design enables the model to condition generation on both acoustic and contextual cues without modifying the underlying pretrained components.

We conduct extensive experiments on the official dataset for the  Interspeech2025 \textbf{M}ulti\textbf{L}ingual \textbf{C}onversational \textbf{S}peech \textbf{L}anguage \textbf{M}odels (MLC-SLM) challenge \citep{mu2025summary}, a large-scale multilingual conversational dataset spanning 11 languages and over 1,500 hours of real-world speech. The experimental results suggest that incorporating context consistently enhances transcription quality, validating the benefits of context-aware generation across diverse linguistic conditions. Our findings further highlight the importance of aligning contextual and acoustic representations, showing that contrastive learning offers additional improvements on different types of context. These results underscore the need for more principled speech–context integration approaches in multilingual ASR.

Our main contributions are as follows: 

\begin{itemize}
    \item We introduce a context-aware SpeechLLM framework for multilingual ASR that effectively harnesses contextual inputs, including dialogue history and biasing words, to enable efficient adaptation across diverse languages, while maintaining a lightweight design.
    
    \item We propose an embedding-level speech context alignment strategy based on contrastive learning, explicitly linking speech features with contextual information to improve semantic grounding in multilingual scenarios.

    \item We conduct comprehensive experiments on a 1,500-hour multilingual dataset across various context settings. Results show consistent improvements over non-contextual and non-contrastive settings, achieving over a 5\% overall performance gain, and provide insights into how contrastive alignment interacts with different context types in multilingual settings.
\end{itemize}

\section{Related Work}

\paragraph{Multilingual ASR}
Recent work in ASR has increasingly focused on multilingual settings, where a single model needs to support many languages, dialects, or accents. For example, \citet{babu2021xls} introduces XLS‑R, a self‑supervised model trained on nearly half a million hours of speech across 128 languages, based on the wav2vec2 \citep{baevski2020wav2vec}. Their evaluation covers both high‑ and low‑resource languages, achieving large error‑rate reductions relative to prior monolingual models.
Similarly, \citet{khurana2022samu} propose SAMU‑XLSR, an utterance‑level multimodal multilingual speech representation model that aligns speech and text embedding spaces across languages by combining XLS‑R and multilingual text embeddings. Other multilingual ASR works extend the unified modeling across many languages, including low‑resource ones \citep{li2025multilingual}. These efforts highlight the value of shared multilingual representation and backbone models for multilingual ASR tasks. 

\paragraph{Context-aware ASR}
Injecting external context, such as preceding utterances, bias‑lists of rare words, or domain‑specific vocabulary, has been shown to improve ASR performance in particular scenarios, such as rare words, proper nouns, and conversational settings \citep{concina2025eloquence, linke2025context}. For instance, \citet{chang2021context} propose Context‑Aware Transformer Transducer, a Transformer‑Transducer architecture that attends over preceding utterances encoded via pretrained BERT or BiLSTM, enabling multi‑turn dialogue context to influence recognition. \citet{huang2020class} introduces Class‑LM \& Word Mapping for contextual biasing in end‑to‑end ASR, allowing a beam search to traverse into a context FST comprised of rare or domain‑specific vocabulary, thus improving recognition of named entities. Beyond these, \citet{fu2023robust} presents Robust Acoustic and Semantic Contextual Biasing, where attention‑based modules incorporate both acoustic and semantic context cues for rare words in neural transducers. \citet{gong2024contextual} discusses Contextual Biasing Speech Recognition in dynamic settings, injecting contexts into earlier encoder layers and assessing runtime cost. 
    
\paragraph{Speech LLM}
With the proliferation of LLMs, recent research has started to bridge speech encoders and LLMs for tasks such as ASR, speech translation, or spoken language understanding. For example, \citet{fan2025alignformer} proposes AlignFormer, a neural adapter connecting a frozen speech encoder to a frozen instruction‑following LLM. Their method uses CTC and dynamic‑window QFormer layers to align heterogeneous modality lengths and preserve the LLM’s instruction‑following capabilities. \citet{chen2024salm} present SALM, which integrates a frozen LLM, audio encoder, and LoRA layers for speech recognition and speech translation with an in‑context learning capability.  \citet{hono2024integrating} proposes a bridge network that maps speech encoder outputs into the LLM embedding space, compressing the sequence length for efficiency. Other work explores end‑to‑end architectures combining pretrained speech encoders with LLMs for ASR \citep{luu2025end}.  These contributions demonstrate the viability of freezing large backbones and training lightweight modules to connect modalities for ASR.

\section{Methodology}

\subsection{Problem Formulation}

Given a spoken utterance $s_t^j$ from dialogue $j$ at turn $t$, along with context information $\mathcal{P}_{ctx_t^j}$, the objective is to generate a textual transcription
$y_t^j = \{ y_1, y_2, \dots, y_L \}$, where $L$ is the output length, that corresponds to the spoken content, considering both the audio signal and the available contextual information.

Formally, we aim to model the conditional probability distribution:
\begin{equation}
    P(y_t^j \mid s_t^j, \mathcal{P}_{ctx_t^j}) = \prod_{l=1}^{L} P(y_l \mid y_{<l}, s_t^j, \mathcal{P}_{ctx_t^j}) ,
\end{equation}

where $y_l$ denotes the $l$-th token in the output sequence, and $y_{<l}$ represents all previously generated tokens.

\subsection{Overall Structure}

\begin{figure*}
    \centering
    \includegraphics[width=0.8\linewidth,height= 0.45\linewidth]{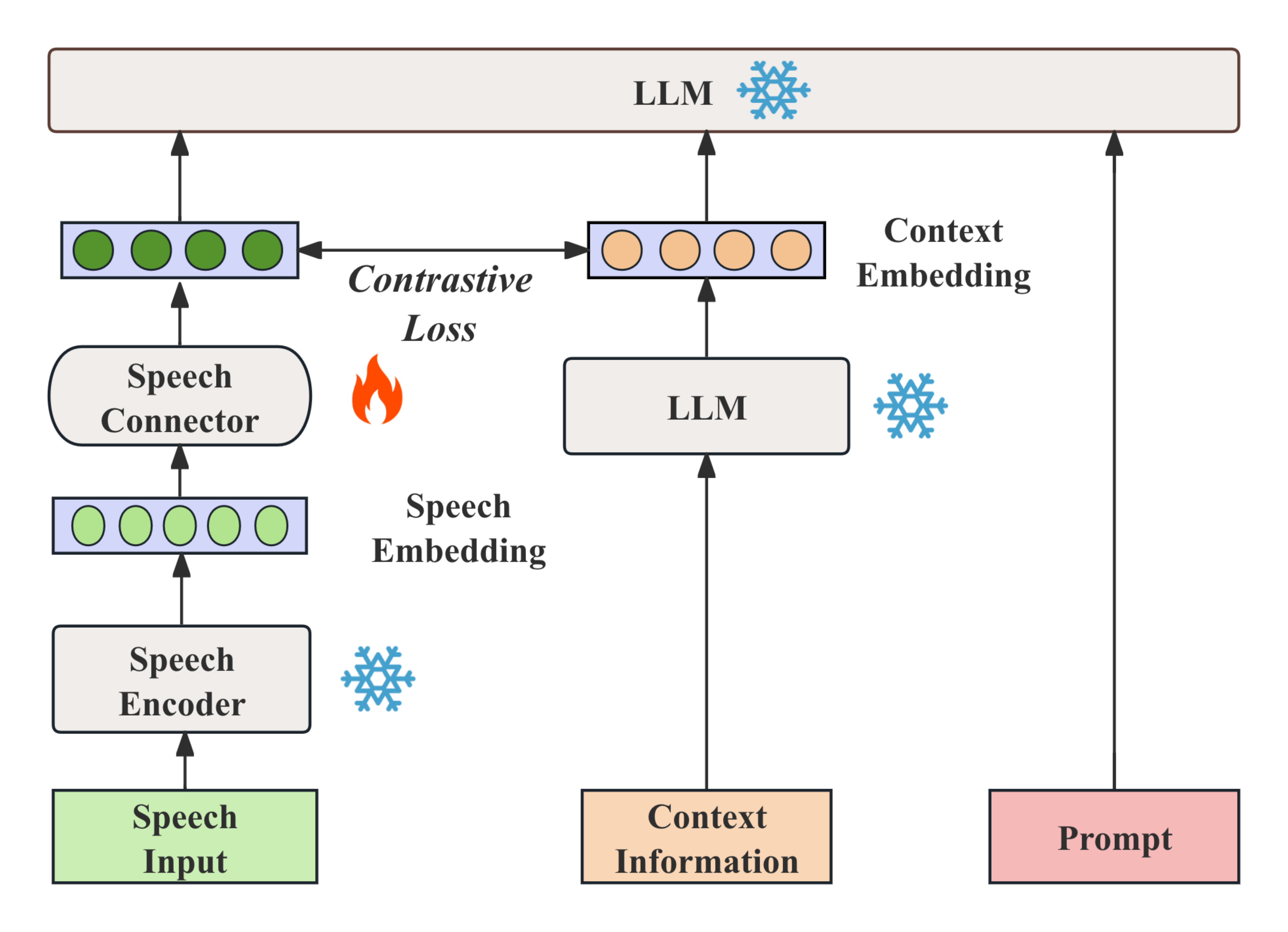}
    \caption{The overall structure of the proposed context-aware multilingual ASR.}
    \label{fig:framework}
\end{figure*}

The overall structure of the proposed context-aware multilingual ASR system is illustrated in Figure~\ref{fig:framework}. Our model architecture integrates a frozen speech encoder, a frozen decoder-only LLM, and a lightweight projection module that bridges the two modalities. To support context-aware transcription, we first construct a model input by injecting contextual information into a fixed instruction template. This prompt and the speech embedding are passed to the LLM, which autoregressively generates the transcription.

The overall process consists of three main stages:  
(1) extracting contextual information relevant to the input speech (Section~\ref{sec:context_extraction});  
(2) projecting the high-dimensional speech features into the shared embedding space of the LLM and aligning them with the contextual prompt. This is achieved through a trainable speech connector (Section~\ref{sec:speech_connector}) and a contrastive learning objective (Section~\ref{sec:contrastive_alignment}) that encourages semantically related speech–context pairs to be close in the representation space;  
(3) generating the final transcription conditioned on both the speech input and the injected context using the LLM decoder.

\subsection{Context Extraction} \label{sec:context_extraction}

In this work, we focus on two types of contextual information to support multilingual ASR: dialogue history and biasing words. Dialogue history refers to the preceding utterances within the same conversation and serves to ground the interpretation of the current utterance in the prior context. This is particularly beneficial for resolving incomplete phrases, pronouns, and context-dependent expressions whose meaning depends on earlier dialogue turns.

In contrast, biasing words refer to keywords or phrases provided in advance, such as named entities, domain-specific terminology, or task-relevant keywords. These terms may be rare or unseen in the training data, but are essential for correctly recognizing domain-relevant content. By jointly incorporating dynamic context (dialogue history) and static prior knowledge (biasing words), we aim to improve the model’s capacity to process complex multilingual speech with greater accuracy and robustness.

\subsubsection{Dialogue History}

To incorporate contextual information, we define the dialogue history of each utterance as the sequence of preceding turns within the same dialogue. 

Let the dataset $\mathcal{D}$ consist of a set of $M$ dialogues: $\mathcal{D} = \{\, \mathcal{D}_i \mid i = 1,2,\dots, M \,\}$, 
where each dialogue $\mathcal{D}_j = \{\, s^j_t \mid t = 1,2,\dots, N_j \,\} (j \in M) $ consists of an ordered sequence of $N_j$ pieces of speech utterances ($N_j$ may vary across dialogues). 
$s^{j}_t$ denotes the $t$-th utterance in $\mathcal{D}_j$ and its corresponding ground-truth transcription is denoted as $y^{j}_t$. 

For a given utterance $s^{j}_t$ in dialogue $\mathcal{D}_j$, we define its dialogue history window of size $K_{DH}$ as $\mathcal{DH}^j_t$:

\begin{equation}
\mathcal{DH}^j_t = 
\begin{cases}
\{ y^{j}_{t-K_{DH}}, \dots, y^{j}_{t-1} \}, & if \quad t > K_{DH} \\
\{ y^{j}_1, \dots, y^{j}_{t-1} \}, & if \quad 1 < t \leq K_{DH} \\
\varnothing, & if \quad t = 1  
\end{cases} .
\end{equation}

To integrate the extracted dialogue history into the model input, we convert $\mathcal{DH}^j_t$ into a natural language prompt. 
When $\mathcal{DH}^j_t \neq \varnothing$, we construct the Dialogue History Prompt $\mathcal{P}_{DH^j_t}$ as: 

"\texttt{The previous} $|\mathcal{DH}^j_t|$ \texttt{turn(s) of this speech is:} $\mathcal{DH}^j_t $."

Here $|\mathcal{DH}^j_t|$ denotes the number of previous utterances available (up to $K_{DH}$), and the entries in $\mathcal{DH}^j_t$ are concatenated with a separator token [SEP]. 

If no conversation history is available (i.e., $t$ = 1), the Dialogue History Prompt $\mathcal{P}_{DH^j_t}$ is defined as :

"\texttt{There is no conversation history of this speech.}"

It should be noted that during training, the dialogue history of each utterance is extracted from the ground-truth transcriptions.
During inference, since ground-truth transcripts are unavailable, we rely on coarse transcriptions generated by a pre-trained multilingual Connectionist Temporal Classification (CTC) model \citep{lugosch2021pseudo}. The same history extraction and prompt formatting procedure used during training is applied to these transcriptions to construct the dialogue history for the test data.

\subsubsection{Biasing Words}

To improve contextual awareness and provide the model with additional lexical cues, we incorporate biasing words, which consist of two categories: (1) Hotwords, short n-gram phrases extracted directly from transcriptions, and (2) Distractor Terms, rare words sampled from a predefined lexicon. These words are included as part of the decoding prompt to highlight potentially informative or underrepresented content, while also encouraging the model to develop robustness against irrelevant lexical cues.

The training hotwords are randomly selected from the transcriptions of speech, following a strategy similar to that used in prior studies \citep{pundak2018deep, huang2023contextualized, futami2024phoneme, yang2024ctc}.
For a given utterance $s^{j}_t$, let its ground-truth transcription be denoted by $y^{j}_t$. To construct the corresponding hotword set $\mathcal{HW}^j_t$, we first tokenize $y^{j}_t$ into a sequence of words. From this sequence, we randomly sample several word-level n-gram phrases. The number of phrases to sample is drawn uniformly from $[1, K_{HW}]$, while the length of each phrase is drawn uniformly from $[1, L_{HW}]$, where $K_{HW}$ and $L_{HW}$ denote the maximum number of phrases and the maximum phrase length, respectively.

Additionally, for each utterance $s^{j}_t$, we include $K_{DT}$ distractor terms $\mathcal{DT}^j_t$, sampled from a predefined, language-specific rare-word lexicon. The rare-word lexicon is constructed following a method similar to that of \citet{le2021contextualized}. Specifically, we construct the rare word lexicon for each language by first aggregating all ground-truth transcriptions in the training set. Unigram frequencies are computed over the tokenized corpus, and words occurring fewer than a threshold $\theta_{rare}$ times are discarded to reduce noise from typos or annotation errors. From the remaining set of words, the bottom $p_{rare}$ fraction, ranked by unigram frequency, is selected to construct the rare-word lexicon.
For each utterance $s^{j}_t$,  the distractor terms $\mathcal{DT}^j_t$ are then sampled from the built rare-word lexicon under the constraint that they do not appear in the current transcription $y^j_t$,

The final biasing words for utterance $s^{j}_t$ is formed by combining the hotwords $\mathcal{HW}^j_t$ and distractor terms $\mathcal{DT}^j_t$. We then convert the biasing words into a Biasing Words Prompt $\mathcal{P}_{BW^j_t}$ :

"\texttt{The speech might contain following words:} $\mathcal{HW}^j_t$, $\mathcal{DT}^j_t$."

Similar to dialogue history extraction, the source of hotwords differs between training and inference. During training, hotwords are derived directly from the ground-truth transcriptions. Since ground-truth is unavailable during inference, hotwords are extracted from the same coarse transcriptions used for dialogue history extraction. The same sampling procedure and prompt formatting are applied in both cases to ensure consistency across training and inference.

\subsection{Speech Connector} \label{sec:speech_connector}

 Given an input speech $s^{j}_t$ , the audio encoder extracts a sequence of high-dimensional acoustic embeddings:

\begin{equation}
    \mathbf{H}_{{spe}_t^j}^{raw} = \mathcal{E}(s^{j}_t) \in \mathbb{R}^{B \times T \times E_a},
\end{equation}

where $\mathcal{E}$ denotes the audio encoder, and $B$, $T$, and $E_a$ denote the batch size, 
the number of acoustic frames, and the audio encoder hidden dimension, respectively.

To align the encoder outputs with the embedding space of the LLM, the speech representations $\mathbf{H}_{{spe}_t^j}^{raw}$ are first downsampled by a factor $K_{down}$, where every $K_{down}$ consecutive frames are concatenated into a single vector. The stacked features are then transformed through two successive linear layers to get the final representation of speech $\mathbf{H}_{{spe}_t^j}$:
\begin{align}
\mathbf{H}_{{spe}_t^j} &= \mathcal{L}_2\!\left( \sigma_{\textsc{gelu}}\!\big(\mathcal{L}_1(\mathbf{H}_{{spe}_t^j}^{stacked})\big) \right), \\
\mathbf{H}_{{spe}_t^j}^{stacked} &= \mathcal{D}_K(\mathbf{H}_{{spe}_t^j}^{raw}) 
\in \mathbb{R}^{B \times \tfrac{T}{K_{down}} \times ( E_a \cdot K_{down})}, 
\end{align}
where $\mathcal{D}_K(\cdot)$ denotes the downsampling operator, $\mathcal{L}_1$ and $\mathcal{L}_2$ denote linear projector, and $\sigma_{\textsc{gelu}}(\cdot)$ is the Gaussian Error Linear Unit (GELU) activation function.

\subsection{Speech–Context Alignment} \label{sec:contrastive_alignment}

To enhance alignment between speech and its associated contextual information, we introduce a contrastive learning objective that encourages paired speech and context embeddings to be close in representation space, while pushing apart mismatched pairs. 

For a given utterance $s^{j}_t$, let the contextual prompt $\mathcal{P}_{ctx^{j}_t}$ (e.g., dialogue history or biasing words) be tokenized and passed through the input embedding layer of the frozen LLM, denoted by $f_{\text{emb}}$. The corresponding context embedding $\mathbf{H}_{ctx^{j}_t} \in \mathbb{R}^{E_t}$ is computed as:

\begin{equation}
    \mathbf{H}_{ctx^{j}_t} = f_{emb}(\mathcal{P}_{ctx^{j}_t}) \in \mathbb{R}^{L \times E_t} , 
\end{equation}

where $L$ is the token length of the context prompt and $E_t$ is the LLM embedding dimension. 

To apply contrastive learning with context embedding and speech embedding, we first conduct mean pooling and then L2-normalization to the context embeding:
\begin{equation}
    \tilde{\mathbf{H}}_{ctx^{j}_t} = \psi\left( \phi \left( \mathbf{H}_{ctx^{j}_t} \right) \right),
\end{equation}
where $\phi(\cdot)$ denotes mean pooling and $\psi(\cdot)$ denotes the L2-normalization.

Similarly, the projected speech embedding $\mathbf{H}_{\text{proj}}$, which is also aggregated and normalized:
\begin{equation}
    \tilde{\mathbf{H}}_{spe^{j}_t} = \psi\left( \phi \left( \mathbf{H}_{spe^{j}_t} \right) \right) .
\end{equation}

For each training batch of size $B$, we define positive pairs as a speech utterance and its corresponding context, and negative pairs as the same speech embedding paired with non‑matching contexts from other utterances in the batch. We then compute the pairwise similarity matrix $\mathbf{S} \in \mathbb{R}^{B \times B}$ using scaled dot products between the normalized speech and context embeddings:

\begin{equation}
S_{t,q} = \frac{ \tilde{\mathbf{H}}_{spe^{j}_t} \cdot \tilde{\mathbf{H}}_{ctx^{k}_q} }{ \tau } ,
\end{equation}

where $s^j_t$ and $s^k_q$ denote two utterances in the batch, $\tilde{\mathbf{H}}_{spe^{j}_t}$ and $\tilde{\mathbf{H}}_{ctx^{k}_q}$ are their corresponding normalized speech and context embeddings, and $\tau$ is a temperature scaling hyperparameter.

The InfoNCE contrastive loss is then computed as:

\begin{equation}
\mathcal{L}_{\text{CL}} = -\frac{1}{B} \sum_{q=1}^{B} \log \frac{ \exp(S_{t,t}) }{ \sum_{q=1}^{B} \exp(S_{t,q}) },
\end{equation}

where each term compares the similarity between the speech embedding $\tilde{\mathbf{H}}_{{spe}^{j}_t}$ and its corresponding context embedding $\tilde{\mathbf{H}}_{{ctx}^{j}_t}$ (i.e., the positive pair) against the similarities between that same speech embedding and all other context embeddings in the batch ${\tilde{\mathbf{H}}_{{ctx}^{k}_q}}$(i.e., the in-batch negatives). This encourages the model to assign higher similarity scores to matched speech–context pairs and lower scores to mismatched ones, thereby learning more discriminative and contextually grounded representations.

\subsection{Training Objective}

For a given utterance $s_t^j$, the decoder outputs a sequence of logits $\mathbf{Z}_t^j \in \mathbb{R}^{L \times V}$, where $L$ is the output length and $V$ is the vocabulary size.
Let the ground-truth transcription be $y_t^j = \{ y_l \}_{l=1}^L, \quad y_l \in [V]$, where $[V] = \{1,2,\dots, V\}$ denotes the index set of the vocabulary, and each $y_l$ is the index of the correct token in this set.

At each position $l$, the probability assigned to the ground-truth index $y_l$ is calculated by the softmax:

\begin{equation}
    p(y_l \mid y_{<l}, s_t^j) = \frac{\exp\!\big(\mathbf{Z}_t^j[l, y_l]\big)}{\sum_{v=1}^V \exp\!\big(\mathbf{Z}_t^j[l, v]\big)},
\end{equation}

where $\mathbf{Z}_t^{\,j}[l,v]$ is the logit for vocabulary index $v$ at position $l$.

The CE loss is then computed as the average negative log-likelihood over all positions:
\begin{equation}
    \mathcal{L}_{CE} = -\frac{1}{L} \sum_{l=1}^L \log p(y_l \mid y_{<l}, s_t^j).
\end{equation}

Finally, the total loss is computed as a weighted combination of CE and CL objectives:

\begin{equation}
\mathcal{L} = \beta \cdot \mathcal{L}_{{CE}} + \alpha \cdot \mathcal{L}_{{CL}},
\end{equation}

where, $\beta$ is a fixed hyperparameter, and $\alpha$ is dynamically adjusted to balance the two losses:

\begin{equation}
\alpha = \frac{ \mathcal{L}_{{CL}} }{ \mathcal{L}_{{CE}} + \mathcal{L}_{{CL}}}.
\end{equation}

The combined objective encourages the decoder to generate accurate transcriptions while enforcing consistency between the speech and contextual representations in the embedding space.

\section{Experiment}

\subsection{Dataset}

We use the MLC-SLM dataset \citep{mu2025summary} for the ASR task to demonstrate the effectiveness of our proposed model.
The selected dataset contains approximately 1571 hours of multilingual conversational data in total, including 1,507 hours for training (Train), 32 hours for validation (Val), and 32 hours for testing (Test). 
Each subset comprises 11 languages: English, French, German, Italian, Portuguese, Spanish, Japanese, Korean, Russian, Thai, and Vietnamese. The English subsets contain accents from various regions: American, Australian, British, Filipino, and Indian. Each recording contains a multi-turn conversational speech of around 20 minutes between two speakers on a randomly assigned topic, including celebrities, dreams, education, emotion, fashion, food, games, the Internet, movies, shopping, travel, etc.

\begin{table}[ht]
\centering
\caption{Durations (hours) of the MLC-SLM dataset.}
\begin{tabular}{lccc}
\hline
\textbf{Language} & \textbf{Train} & \textbf{Val} & \textbf{Test} \\
\hline
English-American     & 100.60 & 2.22 & 2.01 \\
English-Australian   & 100.39 & 2.34 & 2.43 \\
English-British      & 100.48 & 2.23 & 2.03 \\
English-Filipino     & 100.36 & 2.09 & 2.02 \\
English-Indian       & 100.45 & 2.17 & 2.31 \\
French               & 100.38 & 2.26 & 2.07 \\
German               & 100.58 & 2.03 & 2.05 \\
Italian              & 100.67 & 2.10 & 2.19 \\
Japanese             & 100.44 & 2.08 & 2.19 \\
Korean               & 100.68 & 2.03 & 2.02 \\
Portuguese           & 100.33 & 2.18 & 2.14 \\
Russian              & 100.41 & 2.05 & 2.18 \\
Spanish              & 100.47 & 2.14 & 2.18 \\
Thai                 & 100.50 & 2.12 & 2.17 \\
Vietnamese           & 100.45 & 2.15 & 2.01 \\
\hline
\textbf{Total}       & 1507.22 & 32.18 & 32.19 \\
\hline
\end{tabular}
\label{tab:mlc-sml-subsets}
\end{table}

\subsection{Experimental Setting}

\paragraph{General Setting}
In this study, we employ Whisper-large-v3 Turbo \citep{radford2022whisper} as the speech encoder and adopt EuroLLM-1.7B-Instruct \citep{martins2025eurollm} as a LLM decoder. 

During training, the data input is constructed using a structured textual template designed to guide the LLM. Each input sequence takes the form:

\texttt{"<SPEECH> USER: <PROMPT> ASSISTANT: <TRANSCRIPTION>."}

Here, \texttt{<SPEECH>} denotes the speech embedding generated by the speech encoder. \texttt{<TRANSCRIPTION>} is the ground-truth transcription of the speech. \texttt{<PROMPT>} is prefixed task related prompt:

\texttt{"Transcribe the speech to text. The following context information might help:<CONTEXT>"}

In our experiments, we define the \texttt{<CONTEXT>} in three configurations. When only dialogue history is used, \texttt{<CONTEXT>} is set to $\mathcal{P}_{DH}$. When only biasing words are provided, it becomes $\mathcal{P}_{BW}$. When incorporating both dialogue history and biasing words, the context is constructed by concatenating the two: $\mathcal{P}_{DH} + \mathcal{P}_{BW}$.

We also examine the situation where context information is unavailable. In this setting, the <PROMPT> is set as: \texttt{“Transcribe the speech to text.”}

During the inference process, the format of the data
input for the LLM is \texttt{"<SPEECH> USER: <PROMPT> ASSISTANT:"}. 

\paragraph{Model Configuration}
The parameters of both the speech encoder and the LLM are frozen. Optimization is applied only to the lightweight projection module responsible for aligning the speech and language modalities. The model is trained for 2 epochs with a batch size of 8 using the AdamW optimizer, a learning rate of 1e-4, and a weight decay of 1e-6. A linear warm-up schedule with 1,000 warm-up steps is employed to gradually increase the learning rate at the early stage of training. During inference, beam search is used for decoding, with the beam size set to 2.

For context-related hyperparameters, the dialogue history window size $K_{DH}$ is set to 1. The maximum number of hotwords per utterance $K_{HW}$ is 3, and the maximum token length per hotword $L_{HW}$ is 3. For each utterance, $K_{DT}$ = 1 distractor term is introduced. A frequency threshold $\theta_{rare}$ = 2 is used to discard uncommon terms, and the bottom $p_{rare}$ = 10\% of tokens are selected to form the rare-word lexicon. The speech features are downsampled by a factor of $K_{down}$ = 4. The temperature scaling parameter $\tau$ for contrastive learning is set to 0.07.

\begin{table*}[!htbp]
\centering
\caption{WER/CER (\%) results across different contexts (lower is better). Values in brackets indicate the change compared to the corresponding non-CL setting. All = Dialogue History + Biasing Words. BW: Biasing Words; CL: Contrastive Learning. The best results in \textbf{Bold} and the second best \underline{Underlined}}
\resizebox{\linewidth}{!}{
\label{tab:wer_cer_results}
\begin{tabular}{lrrrrrrrc}
\toprule
\textbf{language} & \textbf{No Context} & \textbf{History} & \textbf{BW} & \textbf{All} & \textbf{History+CL} & \textbf{BW+CL} & \textbf{All+CL} & \textbf{metric} \\
\midrule
English-American   & 13.29 & \underline{9.21} & 10.14 & 9.42 & 9.70 (+0.49) & \textbf{9.12} (-1.02) & 10.05 (+0.63) & WER \\
English-Australian & 12.66 & 7.93 & 8.73 & 8.12 & \textbf{7.39} (-0.54) & 7.78 (-0.95) & \underline{7.77} (-0.35) & WER \\
English-British    & 8.58 & 7.46 & 5.96 & \underline{5.78} & \textbf{5.69} (-1.77) & 5.85 (-0.11) & 6.22 (+0.44) & WER \\
English-Filipino   & 10.74 & \underline{9.27} & 10.49 & 11.56 & \textbf{9.15} (-0.12) & 11.69 (+1.20) & 9.88 (-1.68) & WER \\
English-Indian     & 15.91 & 9.97 & \underline{8.43} & \textbf{8.18} & 9.59 (-0.38) & 8.92 (+0.49) & \underline{8.39} (+0.21) & WER \\
French             & \underline{23.32} & 24.06 & 27.00 & 25.01 & \textbf{22.50} (-1.56) & 23.79 (-3.21) & 26.10 (+1.09) & WER \\
German             & 31.49 & \underline{19.89} & 26.14 & 21.99 & \textbf{19.36} (-0.53) & 20.32 (-5.82) & 20.12 (-1.87) & WER \\
Italian            & \underline{20.52} & 25.69 & 21.78 & 20.65 & 20.88 (-4.81) & \underline{20.04} (-1.74) & \textbf{19.87} (-0.78) & WER \\
Japanese           & 38.45 & 25.65 & \textbf{19.43} & 20.96 & 21.27 (-4.38) & 21.74 (+2.31) & \underline{20.60} (-0.36) & CER \\
Korean             & 18.15 & 8.91 & \underline{7.67} & 7.73 & \textbf{7.41} (-1.50) & 8.24 (+0.57) & 7.74 (+0.01) & CER \\
Portuguese         & 44.27 & 32.09 & 32.78 & \underline{31.32} & 36.66 (+4.57) & 35.12 (+2.34) & \textbf{29.61} (-1.71) & WER \\
Russian            & \textbf{16.88} & 20.16 & 19.69 & 20.65 & \underline{18.45} (-1.71) & 19.50 (-0.19) & 18.90 (-1.75) & WER \\
Spanish            & 12.62 & 12.39 & 13.79 & 13.34 & \textbf{10.33} (-2.06) & 11.42 (-2.37) & \underline{11.28} (-2.06) & WER \\
Thai               & 22.73 & \underline{22.45} & 22.58 & 22.78 & \textbf{20.50} (-1.95) & 23.32 (+0.74) & 23.23 (+0.45) & CER \\
Vietnamese         & 25.84 & 12.50 & \textbf{12.26} & 13.69 & \underline{12.35} (-0.15) & 13.78 (+1.52) & 13.80 (+0.11) & WER \\
\midrule
Avg.               & 21.03 & 16.58 & 16.52 & 16.08 & \textbf{15.42} (-1.16) & 16.04 (-0.48) & \underline{15.57} (-0.51) & - \\
\bottomrule
\end{tabular}}
\end{table*}

\paragraph{Evaluation Metrics}

We evaluate model performance using two standard metrics in ASR: Word Error Rate (WER) and Character Error Rate (CER). 

WER and CER are computed at the token level by aligning the predicted transcription with the ground truth using minimum edit distance. The error rate is calculated as:

\begin{equation}
    Error Rate = \frac{S + D + I}{N} ,
\end{equation}

where $S$ is the number of substitutions, $D$ is deletions, $I$ is insertions, and $N$ is the number of tokens (for WER) or characters (for CER) in the reference.

In our multilingual setup, we adopt CER for languages that lack clear word boundaries, including Japanese, Korean, and Thai. For all other languages, where words are separated by spaces, we report WER for evaluation.

\subsection{Results and Discussion}

This section evaluates the proposed context-aware multilingual ASR model across 15 languages under different contextual configurations. Table~\ref{tab:wer_cer_results} presents the experimental results across different contextual configurations.

\paragraph{Effect of contextual information.}
Table~\ref{tab:wer_cer_results} shows that contextual information consistently improves recognition compared to the no-context baseline. The average error rate drops from 21.03\% to 16.08\% when both dialogue history and biasing words are provided, while each individual context type also yields notable improvements. These results confirm that contextual grounding significantly enhances multilingual ASR performance. Gains are particularly strong in German, Korean, and Portuguese, though the impact of context type varies. For German, dialogue history leads to the largest improvement, reducing WER from 31.49\% to 19.89\%, while biasing words are less effective. In contrast, Korean benefits more from biasing words, with CER dropping from 18.15\% to 7.67\%, compared to 8.91\% with history. Portuguese shows gains across all context settings, with the lowest WER of 29.61\% achieved when both history and biasing are used with contrastive learning.
These results demonstrate the value of leveraging both dynamic dialogue history and static lexical cues, although the effectiveness varies across languages.

\paragraph{Role of contrastive learning.}
The impact of contrastive learning shows a consistent positive trend across all context settings, though the degree of improvement varies. The best-performing setting is dialogue history combined with contrastive learning, which achieves the lowest average error rate of 15.42\%, improving over the history-only setting at 16.58\% by 1.16\%. This indicates that contrastive alignment is especially effective in leveraging conversational history, helping the model maintain semantic coherence and resolve context-dependent expressions more reliably.
When applied to biasing words alone, contrastive learning also yields gains, with the error rate reduced from 16.52\% to 16.04\%. This suggests that aligning speech with lexical context offers some benefits for recognizing rare or domain-specific terms, but further design improvements are needed to fully exploit its potential.
The setting combining both dialogue history and biasing words with contrastive learning produces the second-best performance at 15.57\%. Although this improves upon the non-contrastive setting at 16.08 \%, the gain is smaller than that achieved with history alone. This pattern suggests that merging heterogeneous context types under a single alignment objective may introduce competing signals. While dialogue history emphasizes semantic continuity, biasing words highlight local lexical anchors, and contrastive learning may struggle to reconcile both simultaneously.
Overall, contrastive learning consistently enhances performance across all settings, with the most significant improvements observed when applied to dialogue history alone. These findings underscore the value of targeted contrastive alignment and motivate future work on context-specific or disentangled optimization strategies.

\paragraph{Language-specific behaviors.} 
The results also reveal distinct language-specific patterns. For the English dialects, contextual information consistently improves recognition, with British English achieving the lowest error rate when dialogue history is combined with contrastive learning. French exhibits only modest improvements. While dialogue history with contrastive learning slightly outperforms the baseline, configurations that rely on biasing words often reduce accuracy, indicating that errors in coarse transcriptions or distractor terms can introduce misleading signals. For Italian, dialogue history alone degrades performance, but biasing words with contrastive learning provide small yet consistent gains. Japanese achieves its best performance with biasing words alone; adding contrastive learning degrades results slightly, possibly due to challenges in constructing reliable contrastive negatives for its complex writing system. Portuguese benefits most from combining dialogue history and biasing words with contrastive learning, contrary to expectations of degradation, suggesting alignment enhances contextual understanding. Spanish shows strong improvements when dialogue history is paired with contrastive learning. Russian, however, sees limited gains or mild degradation from contextual cues, indicating that cross-turn prompts are less effective under current settings.

\paragraph{Unseen Languages in Pretraining.}

Thai and Vietnamese, which are not included in EuroLLM pre-training, provide insight intothe challenges of generalization to unseen languages. Vietnamese shows clear gains from contextual information, with WER dropping by roughly half compared to no context setting. Incorporating dialogue history with contrastive learning further stabilizes performance, indicating that cross-turn grounding can transfer effectively even when the target language is unseen during pre-training of the LLM. The results for Thai show variability. While dialogue history and contrastive learning yield minor improvements, integrating all context types with contrastive learning results in a decline. This pattern suggests that in languages characterized by tonal complexity and noisier biasing terms, the simultaneous alignment of multiple context sources can amplify recognition errors rather than mitigate them.

Overall, the results indicate that contextual information substantially improves multilingual ASR, though its effectiveness varies with the type of context and the target language. Contrastive learning strengthens the use of dialogue history but becomes less stable when heterogeneous contexts are introduced. The contrasting results show that effective context-aware ASR requires careful control of biasing terms and prompt design to avoid instability and preserve the benefits of contextual modeling.

\section{Conclusion}

This paper introduces a multilingual ASR framework designed to integrate contextual information into speech recognition without modifying the underlying speech encoder or language model. By combining a frozen speech encoder, a decoder-only LLM, and a lightweight projection module, the system supports various types of contextual input, such as dialogue history and biasing words, in a modular and efficient way.

We evaluate the proposed model on a real-world dataset, covering 11 different languages and multiple English accents. 
The results show that contextual information consistently improves recognition across all languages and accents, with an average error rate reduction of over 5\% compared to the no-context baseline. Contrastive learning further enhances performance when applied to different context types, especially dialogue history, which achieves the best overall results. For instance, the combination of history and contrastive learning yields the lowest average error rate across settings, with substantial improvements in German, Korean, and Portuguese. Surprisingly, combining both dialogue history and biasing words with contrastive learning does not lead to the best performance and in some cases slightly underperforms the all context-only setting. This suggests that while contextual cues and alignment objectives are beneficial, their interaction in multi-context setups can introduce interference, highlighting the need for more adaptive integration strategies.

\section{Ethics Statements}

This work focuses on improving multilingual ASR through context-aware generation and alignment between speech and language modalities. All experiments were conducted using the publicly available MLC-SLM dataset containing real-world, consented multilingual speech data spanning diverse languages and dialects. We ensured that no personally identifiable information or sensitive content was used in training or evaluation. Additionally, care should be taken when deploying the system in high-stakes applications, where misrecognition or inappropriate use of contextual prompts could lead to biased or misleading outputs.

\section{Limitations}

This work investigates context-aware multilingual ASR using two specific types of context: dialogue history and biasing words. While these choices cover common conversational and domain-specific cues, other potentially useful context signals, such as speaker identity, acoustic environment, or visual grounding, are not explored and may further improve transcription quality. Additionally, our evaluation is limited to the MLC-SLM dataset using EuroLLM. Generalization to under-represented languages, out-of-domain distributions, and low-resource or noisy conditions remains an open question. Future work could explore broader contextual types and extend evaluation to additional pretrained models and multilingual datasets.

\section{Acknowledgments}
This work has been partially funded by the European Union’s Horizon Europe research and innovation programme under the project ELOQUENCE (Grant Agreement No. 101135916) and the UKRI. Views and opinions expressed are, however, those of the author(s) only and do not necessarily reflect those of the European Union or the Research Executive Agency. Neither the European Union nor the granting authority can be held responsible for them.

\section{Bibliographical References}\label{sec:reference}
\bibliographystyle{lrec2026-natbib}
\bibliography{ref}

\end{document}